\documentclass[runningheads]{llncs}
\usepackage[T1]{fontenc}

\usepackage{graphicx}
\usepackage{arydshln}
\usepackage{times}
\usepackage{latexsym}
\usepackage{spverbatim}
\usepackage{adjustbox}
\usepackage{fvextra}
\usepackage{amsmath}
\usepackage{lscape}
\usepackage{longtable}
\usepackage{tikz}
\usepackage{graphicx}
\usepackage{booktabs}
\usepackage{algorithm}
\usepackage{algorithmic}
\usepackage{tabularx}
\usepackage{makecell}
\usepackage{setspace}
\usepackage{ragged2e}
\usepackage{tabulary}
\usepackage{enumitem}
\usepackage{rotating}
\usepackage{dblfloatfix}
\usepackage{rotating}
\usepackage{multirow}
\usepackage{pifont}
\usepackage{caption}
\usepackage{booktabs}
\usepackage{colortbl}
\usepackage{hyperref}
\usepackage{xcolor}

\begin{document}
\newcommand{\name}{\textit{Auepora}}
\newcommand{\lname}{A Unified Evaluation Process of RAG}
\newcommand{\modify}[1]{#1}
\title{Evaluation of Retrieval-Augmented Generation: \\ A Survey}
%
%
\author{
    Hao Yu\inst{1,2} \and
    Aoran Gan\inst{3} \and
    Kai Zhang\inst{3} \and
    Shiwei Tong\inst{1}$^{\dagger }$ \and
    Qi Liu\inst{3} \and
    Zhaofeng Liu\inst{1}
}

\institute{
    Tencent Company \and McGill University \and
    State Key Laboratory of Cognitive Intelligence, \\University of Science and Technology of China \\
    \email{hao.yu2@mail.mcgill.ca} \\
    \email{gar@mail.ustc.edu.cn}\\
    \email{\{shiweitong\inst{}$^{\dagger }$,zhaofengliu\}@tencent.com}\\
    \email{\{kkzhang08,qiliuql\}@ustc.edu.cn}
}

\maketitle
\begin{abstract}
\modify{Retrieval-Augmented Generation (RAG) has recently gained traction in natural language processing. Numerous studies and real-world applications are leveraging its ability to enhance generative models through external information retrieval. Evaluating these RAG systems, however, poses unique challenges due to their hybrid structure and reliance on dynamic knowledge sources. To better understand these challenges, we conduct \textit{\lname{}} (\name) and aim to provide a comprehensive overview of the evaluation and benchmarks of RAG systems. Specifically, we examine and compare several quantifiable metrics of the Retrieval and Generation components, such as relevance, accuracy, and faithfulness, within the current RAG benchmarks, encompassing the possible output and ground truth pairs. We then analyze the various datasets and metrics, discuss the limitations of current benchmarks, and suggest potential directions to advance the field of RAG benchmarks.}

\end{abstract}

\def\thefootnote{$\dagger$}\footnotetext{Corresponding Author}\def\thefootnote{\arabic{footnote}}
\def\thefootnote{}\footnotetext{Paper Homepage: \url{https://github.com/YHPeter/Awesome-RAG-Evaluation} }\def\thefootnote{\arabic{footnote}}

\section{\modify{Introduction}}
Retrieval-Augmented Generation (RAG) \cite{Lewis2020} efficiently enhances the performance of generative language models through integrating information retrieval techniques. It addresses a critical challenge faced by standalone generative language models: the tendency to produce responses that, while plausible, may not be grounded in facts. By retrieving relevant information from external sources, RAG significantly reduces the incidence of hallucinations \cite{Huang2023} or factually incorrect outputs, thereby improving the content's reliability and richness. \cite{Zhang2023} This fusion of retrieval and generation capabilities enables the creation of responses that are not only contextually appropriate but also informed by the most current and accurate information available, making RAG a development in the pursuit of more intelligent and versatile language models \cite{Zhang2023,Yao2023}.

\begin{figure}[ht!]
    \centering
    \adjustbox{center}{
        \includegraphics[width=\linewidth]{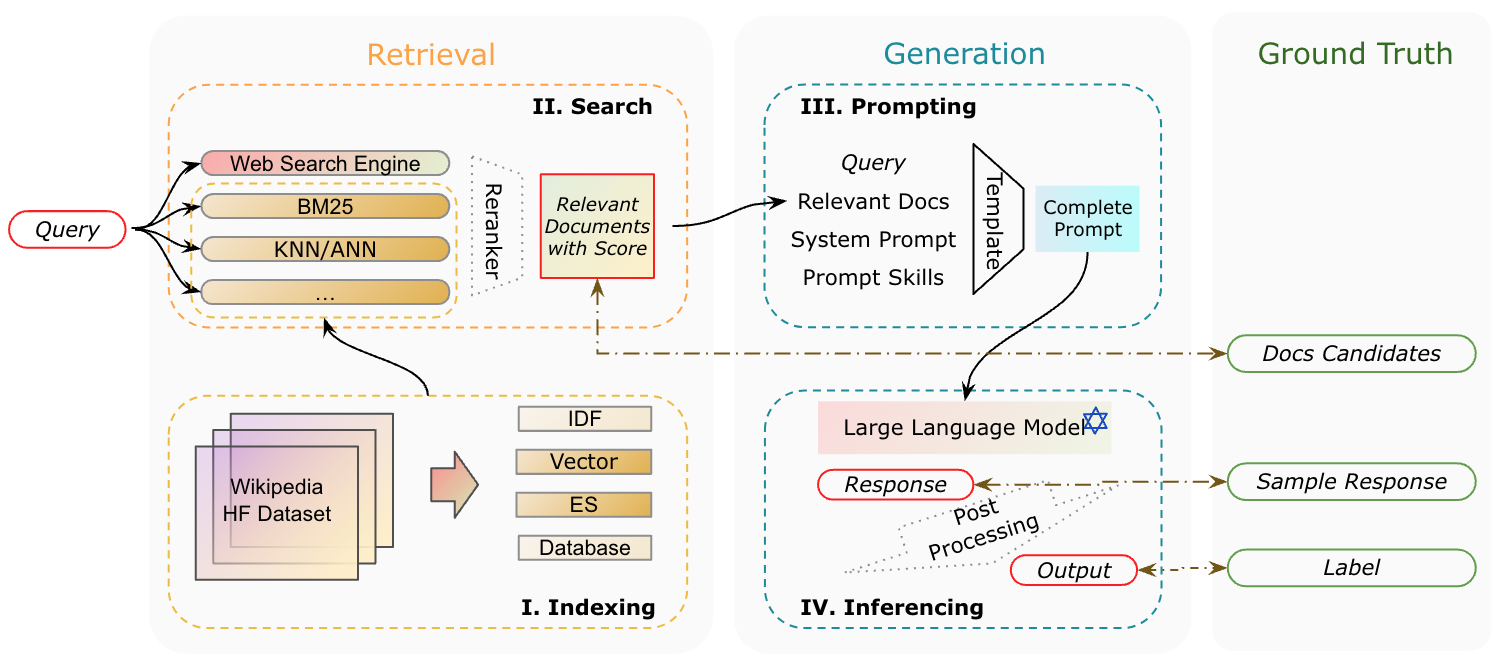}
    }
        \caption{The structure of the RAG system with retrieval and generation components and corresponding four phrases: indexing, search, prompting and inferencing. The pairs of ``Evaluable Outputs'' (EOs) and ``Ground Truths'' (GTs) are highlighted in \textcolor{red}{read frame} and \textcolor[HTML]{9acd32}{green frame}, with \textcolor{brown}{brown dashed arrows}.}
        \label{fig:rag-structure}
        \vspace{-0.2cm}
    \end{figure}

Numerous studies of RAG systems have emerged from various perspectives since the advent of Large Language Models (LLMs) \cite{Transformer,GPT2,Emergent,InstructGPT,GPT4,Zhang2022a,RAGServey}.
The RAG system comprises two primary components: \textit{\textbf{Retrieval}} and \textit{\textbf{Generation}}. 
The retrieval component aims to extract relevant information from various external knowledge sources. It involves two main phases, \textit{indexing} and \textit{searching}. Indexing organizes documents to facilitate efficient retrieval, using either inverted indexes for sparse retrieval or dense vector encoding for dense retrieval \cite{RAGServey,FAISS,ColBERT}. The searching component utilizes these indexes to fetch relevant documents on the user's query, often incorporating the optional rerankers \cite{HaystackDiversity,CRUD,RGB,MultiHop-RAG} to refine the ranking of the retrieved documents.
The generation component utilizes the retrieved content and question query to formulate coherent and contextually relevant responses with the prompting and inferencing phases. As the ``Emerging'' ability \cite{Emergent} of LLMs and the breakthrough in aligning human commands \cite{InstructGPT}, LLMs are the best performance choices model for the generation stage. Prompting methods like Chain of Thought (CoT) \cite{CoT}, Tree of Thgouht \cite{Yao2023a}, Rephrase and Respond (RaR) \cite{RaR} guide better generation results. In the inferencing step, LLMs interpret the prompted input to generate accurate and in-depth responses that align with the query's intent and integrate the extracted information \cite{Lewis2021,Devlin2019} without further finetuning, such as fully finetuning \cite{RAGServey,RAGorFT,Zhang2021,Zhang2019a} or LoRA \cite{LoRA}. Appendix \ref{appendix:rag-structure} details the complete RAG structure.
Figure \ref{fig:rag-structure} illustrates the structure of the RAG systems as mentioned.

The importance of evaluating RAG is increasing in parallel with the advancement of RAG-specific methodologies. On the one hand, RAG is a complex system intricately tied to specific requirements and language models, resulting in various evaluation methods, indicators, and tools, particularly given the black-box LLM generation. Evaluating RAG systems involves specific components and the complexity of the overall system assessment.
On the other hand, the complexity of RAG systems is further compounded by the external dynamic database and the various downstream tasks, such as content creation or open domain question answering \cite{RAGServey,Zhang2023c}. These challenges necessitate the development of comprehensive evaluation metrics that can effectively capture the interplay between retrieval accuracy and generative quality \cite{Barnett2024,Cuconasu2024}.
To clarify the elements further, we try to address the current gaps in the area, which differs from the prior RAG surveys \cite{AIGCSurvey,RAGServey,RATGSurvey} that predominantly collected specific RAG methods or data. We have compiled 12 distinct evaluation frameworks, encompassing a range of aspects of the RAG system. Following the procedure of making benchmarks, we analyze through targets, datasets and metrics mentioned in these benchmarks and summarize them into \textbf{\lname{}} (\textit{\name}) as three corresponding phases.\\

For this paper, we contribute in the following aspects:
\begin{enumerate}
    \item \textbf{Challenge of Evaluation}: This is the first work that summarizes and classifies the challenges in evaluating RAG systems through the structure of RAG systems, including three parts retrieval, generation, and the whole system.
    \item \textbf{Analysis Framework}: In light of the challenges posed by RAG systems, we introduce an analytical framework, referred to as \textit{\lname{}} (\name), which aims to elucidate the unique complexities inherent to RAG systems and guide for readers to comprehend the effectiveness of RAG benchmarks across various dimensions
    \item \textbf{RAG Benchmark Analysis}: With the help of \textit{\name{}}, we comprehensively analyze existing RAG benchmarks, highlighting their strengths and limitations and proposing recommendations for future developments in RAG system evaluation.
\end{enumerate}

\section{\modify{Challenges in Evaluating RAG Systems}}

Evaluating hybrid RAG systems entails evaluating retrieval, generation and the RAG system as a whole. These evaluations are multifaceted, requiring careful consideration and analysis. Each of them encompasses specific difficulties that complicate the development of a comprehensive evaluation framework and benchmarks for RAG systems.

\paragraph{\textbf{Retrieval}}

The retrieval component is critical for fetching relevant information that informs the generation process. One primary challenge is the dynamic and vast nature of potential knowledge bases, ranging from structured databases to the entire web. This vastness requires evaluation metrics that can effectively measure the precision, recall, and relevance of retrieved documents in the context of a given query \cite{MultiHop-RAG,LangChain2023}. Moreover, the temporal aspect of information, where the relevance and accuracy of data can change over time, adds another layer of complexity to the evaluation process \cite{RGB}. Additionally, the diversity of information sources and the possibility of retrieving misleading or low-quality information pose significant challenges in assessing the effectiveness of filtering and selecting the most pertinent information \cite{CRUD}. The traditional evaluation indicators for retrieval, such as Recall and Precision, cannot fully capture the nuances of RAG retrieval systems, necessitating the development of more nuanced and task-specific evaluation metrics \cite{ARES}.

\paragraph{\textbf{Generation}}

The generation component, powered by LLMs, produces coherent and contextually appropriate responses based on the retrieved content. The challenge here lies in evaluating the faithfulness and accuracy of the generated content to the input data. This involves not only assessing the factual correctness of responses but also their relevance to the original query and the coherence of the generated text \cite{LLMasJudge,ARES}. The subjective nature of certain tasks, such as creative content generation or open-ended question answering, further complicates the evaluation, as it introduces variability in what constitutes a ``correct'' or ``high-quality'' response \cite{ResearchyQ}.

\paragraph{\textbf{RAG System as a Whole}}

Evaluating the whole RAG system introduces additional complexities. The interplay between the retrieval and generation components means that the entire system's performance cannot be fully understood by evaluating each component in isolation \cite{ARES,RAGAS}. The system needs to be assessed on its ability to leverage retrieved information effectively to improve response quality, which involves measuring the added value of the retrieval component to the generative process. Furthermore, practical considerations such as response latency and the ability to handle ambiguous or complex queries are also crucial for evaluating the system's overall effectiveness and usability \cite{CRUD,RGB}. 

\paragraph{Conclusion}
Evaluating the target shift from traditional absolute numeric metrics to multi-source and multi-target generation evaluation, along with the intricate interplay between retrieval and generation components, poses significant challenges. \cite{EvalLLM,EvalNLG} Searches in a dynamic database may lead to misleading results or contradict the facts. 
Diverse and comprehensive datasets that accurately reflect real-world scenarios are crucial. 
Challenges also arise in the realm of metrics, encompassing generative evaluation criteria for distinct downstream tasks, human preferences, and practical considerations within the RAG system. 
Most prior benchmarks predominantly tackle one or several aspects of the RAG assessment but lack a comprehensive, holistic analysis.

\section{\modify{\lname{}} (\name)}
To facilitate a deeper understanding of RAG benchmarks, we introduce \textit{\lname{}} (\name{}), which focuses on three key questions of benchmarks: \textit{What to Evaluate? How to Evaluate? How to Measure?} which correlated to \textit{Target}, \textit{Dataset}, and \textit{Metric} respectively.
We aim to provide a clear and accessible way for readers to comprehend the complexities and nuances of RAG benchmarking.

The \textit{Target} module is intended to determine the evaluation direction. The \textit{Dataset} module facilitates the comparison of various data constructions in RAG benchmarks. The final module, \textit{Metrics}, introduces the metrics that correspond to specific targets and datasets used during evaluation.
Overall, it is designed to provide a systematic methodology for assessing the effectiveness of RAG systems across various aspects by covering all possible pairs at the beginning between the ``Evaluable Outputs'' (EOs) and ``Ground Truths'' (GTs). In the following section, we will explain thoroughly \name and utilize it for introducing and comparing the RAG benchmarks.

\begin{figure*}[bhtp]
    \centering
    \includegraphics[width=\linewidth]{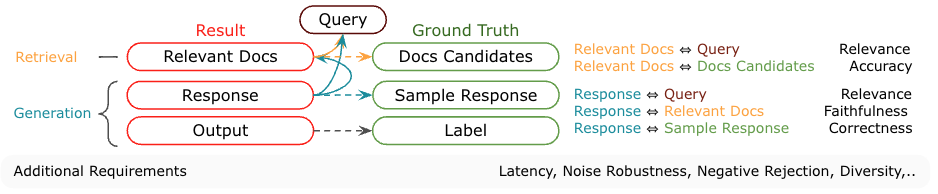}
    \caption{The \textit{Target} modular of the \name.}
    \label{fig:aspect-display}
\vspace{-0.2cm}
\end{figure*}

\subsection{Evaluation Target (\textit{What to Evaluate?})}

The combination of EOs and GTs in the RAG system can generate all possible targets, which is the fundamental concept of the \name{} (as shown in Figure \ref{fig:rag-structure}). Once identified, these targets can be defined based on a specific pair of EOs or EO with GT, as illustrated in Figure \ref{fig:aspect-display}, and used to analyze all aspects of current RAG benchmarks.

\subsubsection{Retrieval}
The EOs are the relevant documents for evaluating the retrieval component depending on the query. Then we can construct two pairwise relationships for the retrieval component, which are \textit{Relevant Documents $\leftrightarrow$ Query}, \textit{Relevant Documents $\leftrightarrow$ Documents Candidates}.
\begin{itemize}
    \item[-] \textbf{Relevance} (\textit{Relevant Documents $\leftrightarrow$ Query}) evaluates how well the retrieved documents match the information needed expressed in the query. It measures the precision and specificity of the retrieval process.
    \item[-] \textbf{Accuracy} (\textit{Relevant Documents $\leftrightarrow$ Documents Candidates}) assesses how accurate the retrieved documents are in comparison to a set of candidate documents. It is a measure of the system's ability to identify and score relevant documents higher than less relevant or irrelevant ones.
\end{itemize}

\begin{table}[hb!]
\centering
\vspace{-1cm}

\setlength{\abovecaptionskip}{0.4cm}
\caption{
The evaluating targets and corresponding metrics across various frameworks for evaluating RAG systems. The presentation distinguishes between the core areas of Retrieval and Generation considered in the evaluation. 
The different aspects of the evaluation are set as different colours in the table:
\textcolor[HTML]{9A0000}{{Relevance}}, \textcolor[HTML]{CD9934}{{Accuracy}} of Retrieval and \textcolor[HTML]{F8A102}{{Faithfulness}}, \textcolor[HTML]{036400}{{Correctness}} and \textcolor[HTML]{6665CD}{{Relevance}} of Generation.
 The consideration of the \textit{Additional Requirements} beyond the retrieval and generation component is also collected.
Noted that quite a few of the works employed multiple methods or evaluated multiple aspects simultaneously. 
}

\resizebox{\textwidth}{!}{
\renewcommand{\arraystretch}{1.5} 

\begin{tabular}{lcc|ccc}
\toprule
Category & Framework & Time  &  \textbf{Raw Targets}&\textbf{Retrieval} & \textbf{Generation}  \\ \hline

Tool & TruEra RAG Triad \cite{TruEra} & 2023.10  & \makecell[c]{\color[HTML]{9A0000}Context Relevance \\ \color[HTML]{6665CD}Answer Relevance \\ \textit{Groundedness}}&\color[HTML]{9A0000} LLM as a Judge & \color[HTML]{6665CD} LLM as a Judge\\\hline
Tool & LangChain Bench. \cite{LangChain2023} & 2023.11   & \makecell[c]{\color[HTML]{CD9934} Accuracy \\ \color[HTML]{F8A102} Faithfulness \\ \textit{Execution Time}\\ \textit{Embed. CosDistance}} &\color[HTML]{CD9934} Accuracy & \color[HTML]{F8A102} LLM as a Judge   \\\hline
Tool & Databricks Eval \cite{DatabricksRAGEval} & 2023.12  & \makecell[c]{{\color[HTML]{036400}Correctness} \\ \textit{Readability} \\ \textit{Comprehensiveness}}
&- & \color[HTML]{036400} LLM as a Judge   \\ \hline
Benchmark & RAGAs \cite{RAGAS} & 2023.09  & \makecell[c]{  \textcolor[HTML]{9A0000}{Context Relevance} \\\textcolor[HTML]{6665CD}{Answer Relevance} \\ \textcolor[HTML]{F8A102}{Faithfulness}}&\color[HTML]{9A0000} LLM as a Judge & \makecell[c]{\color[HTML]{6665CD} LLM Gen + CosSim\\ \color[HTML]{F8A102} LLM as a Judge} \\\hline
Benchmark & RECALL \cite{RECALL} & 2023.11  &  \makecell[c]{{\color[HTML]{036400}Response Quality} \\ \textit{Robustness}} &- & \color[HTML]{036400} BLEU, ROUGE-L  \\\hline
Benchmark & ARES \cite{ARES} & 2023.11  & \makecell[c]{ \textcolor[HTML]{9A0000}{Context Relevance} \\\textcolor[HTML]{F8A102}{Answer Faithfulness} \\ \textcolor[HTML]{6665CD}{Answer Relevance}}&\color[HTML]{9A0000} LLM + Classifier  &\makecell[c]{\color[HTML]{F8A102} LLM + Classifier \\\color[HTML]{6665CD} LLM + Classifier}  \\\hline
Benchmark & RGB \cite{RGB} & 2023.12  & \makecell[c]{\color[HTML]{F8A102} Information Integration \\\textit{Noise Robustness} \\ \textit{Negative Rejection} \\\textit{Counterfactual Robustness}} &- &\color[HTML]{F8A102} Accuracy\\ \hline
Benchmark & MultiHop-RAG \cite{MultiHop-RAG} & 2024.01  & \makecell[c]{ \color[HTML]{CD9934}Retrieval Quality \\ \color[HTML]{036400} Response Correctness}
&\color[HTML]{CD9934} MAP, MRR, Hit@K & \color[HTML]{036400} LLM as a Judge   \\\hline
Benchmark & CRUD-RAG \cite{CRUD} & 2024.02  & \makecell[c]{\textit{CREATE}, {\color[HTML]{036400}RE}{\color[HTML]{F8A102}AD} \\ \textit{UPDATE}, \textit{DELETE}} & - & \makecell[c]{\color[HTML]{036400} ROUGE, BLEU \\ \color[HTML]{F8A102}RAGQuestEval } \\\hline
Benchmark & MedRAG \cite{MedRAGBench} & 2024.02  &  \makecell[c]{\color[HTML]{036400}Accuracy}&- &\color[HTML]{036400} Accuracy  \\\hline
 Benchmark & FeB4RAG \cite{FeB4RAG}& 2024.02  &  \makecell[c]{\textcolor[HTML]{F8A102} {Consistency} \\ \textcolor[HTML]{036400} {Correctness} \\ \textit{Clarity} \\ \textit{Coverage}}& - & \makecell[c]{\textcolor[HTML]{F8A102}{Human Evaluation} \\ \textcolor[HTML]{036400}{Human Evaluation}} \\\hline
 Benchmark & CDQA \cite{CDQA} & 2024.03  &\makecell[c]{\color[HTML]{036400}Accuracy}&- & \color[HTML]{036400} F1  \\\hline
 Benchmark & DomainRAG \cite{DomainRAG} & 2024.06  &\makecell[c]{\textcolor[HTML]{036400}{Correctness} \\\textcolor[HTML]{F8A102}{Faithfulness} \\ \textit{Noise Robustness}\\ \textit{Structural Output}}&- & \makecell[c]{\color[HTML]{036400}  F1, Exact-Match \\ \color[HTML]{F8A102} Rou\color[HTML]{036400}{ge-L} \\ \color[HTML]{F8A102} LLM as a Judge}\\\hline
 Benchmark & ReEval \cite{ReEval} & 2024.06 &\textcolor[HTML]{F8A102}{Halluci}\textcolor[HTML]{036400}{nation}&- & \makecell[c]{\textcolor[HTML]{F8A102}{F1, Exac}\textcolor[HTML]{036400}{ct-Match} \\ \textcolor[HTML]{F8A102}{LLM as }\textcolor[HTML]{036400}{a Judge }\\ \textcolor[HTML]{F8A102}{Human E}\textcolor[HTML]{036400}{valuation}}\\\hline
 Research& FiD-Light \cite{EfficientRAG}& 2023.07& \textit{Latency}&- & -\\ \hline
 Research& Diversity Reranker \cite{HaystackDiversity}& 2023.08& \textit{Diversity}&Cosine Distance & -\\ \bottomrule
\end{tabular}%
}

\label{table:aspect-metrics}
\end{table}

\subsubsection{Generation}
The similar pairwise relations for the generation components are listed below. The EOs are the generated text and phrased structured content. Then we need to compare these EOs with the provided GTs and labels.

\begin{itemize}
    \item[-] \textbf{Relevance} (\textit{Response $\leftrightarrow$ Query}) measures how well the generated response aligns with the intent and content of the initial query. It ensures that the response is related to the query topic and meets the query's specific requirements.
    \item[-] \textbf{Faithfulness} (\textit{Response $\leftrightarrow$ Relevant Documents}) evaluates if the generated response accurately reflects the information contained within the relevant documents and measures the consistency between generated content and the source documents.
    \item[-] \textbf{Correctness} (\textit{Response $\leftrightarrow$ Sample Response}) Similar to the accuracy in the retrieval component, this measures the accuracy of the generated response against a sample response, which serves as a ground truth. It checks if the response is correct in terms of factual information and appropriate in the context of the query.
    
\end{itemize}

The targets of Retrieval and Generation components are introduced. Table \ref{table:aspect-metrics} lists the relative work on improving and evaluating RAG and its benchmarks cut off in June 2024. Table \ref{table:aspect-metrics} portrays this information, where each evaluation criterion is represented by a different colour. For example, FeB4RAG \cite{FeB4RAG}, the fourth from the last, has posited four standards based on \cite{EvalGenAdHocIR} that comprise Consistency, Correctness, Clarity, and Coverage. \textcolor[HTML]{036400}{Correctness} is equivalent to accuracy in retrieval, and \textcolor[HTML]{F8A102}{Consistency} is tantamount to faithfulness in the generation component. While accuracy in retrieval gauges the correctness of the retrieved information, we posit that Coverage pertains to the coverage rate and is more associated with diversity. Therefore, we consider \textit{Coverage} to be linked with diversity and an additional requirement in our proposed evaluation framework, which will be introduced subsequently. The remaining standard, \textit{Clarity}, is also classified as an additional requirement in our proposed framework. The other tools and benchmarks are processed similarly.

Tools and benchmarks offer varying degrees of flexibility in evaluating datasets for RAG systems. Tools, which specify only evaluation targets, provide a versatile framework capable of constructing complete RAG applications and evaluation pipelines, as seen in works like \cite{TruEra,LangChain2023,DatabricksRAGEval}. Benchmarks, on the other hand, focus on different aspects of RAG evaluation with specific emphasis on either retrieval outputs or generation targets. For instance, RAGAs \cite{RAGAS} and ARES \cite{ARES} assess the relevance of retrieval documents, while RGB and MultiHop-RAG \cite{RGB,MultiHop-RAG} prioritize accuracy, necessitating comparison with GTs. The \cite{ReEval} focuses on the Hallucination, which is a combination of faithfulness and correctness. All benchmarks consider generation targets due to their critical role in RAG systems, though their focus areas vary.

\subsubsection{Additional Requirement}
In addition to evaluating the two primary components outlined, a portion of the works also addressed some additional requirements of RAG (Black and \textit{Italics} targets in Table \ref{fig:aspect-display}). The requirements are as follows:

\begin{itemize}
    \item[-] \textbf{Latency} \cite{EfficientRAG,LangChain2023} measures how quickly the system can find information and respond, crucial for user experience.
    \item[-] \textbf{Diversity} \cite{HaystackDiversity,LangChain2023} checks if the system retrieves a variety of relevant documents and generates diverse responses.
    \item[-] \textbf{Noise Robustness} \cite{RGB} assesses how well the system handles irrelevant information without affecting response quality.
    \item[-] \textbf{Negative Rejection} \cite{RGB} gauges the system's ability to refrain from providing a response when the available information is insufficient.
    \item[-] \textbf{Counterfactual Robustness} \cite{RGB} evaluates the system's capacity to identify and disregard incorrect information, even when alerted about potential misinformation.
    \item[-] \textbf{More}: For more human preferences considerations, there can be more additional requirements, such as readability \cite{FeB4RAG,DatabricksRAGEval}, toxicity, perplexity \cite{DatabricksRAGEval}, etc.
\end{itemize}



For the exception, CRUD-RAG \cite{CRUD} introduces a comprehensive benchmark addressing the broader spectrum of RAG applications beyond question-answering, categorized into Create, Read, Update, and Delete scenarios. 
This benchmark evaluates RAG systems across diverse tasks, including text continuation, question answering, hallucination modification, and multi-document summarization. It offers insights for optimizing RAG technology across different scenarios. DomainRAG \cite{DomainRAG} identifies six complex abilities for RAG systems: conversational, structural information, faithfulness, denoising, time-sensitive problem solving, and multi-doc understanding. ReEval \cite{ReEval} specifically targets hallucination evaluation by employing a cost-effective LLM-based framework that utilizes prompt chaining to create dynamic test cases. 


\begin{table}[ht!]
\centering

\setlength{\abovecaptionskip}{0.35cm}
\caption{\small The evaluation datasets used for each benchmark. The dataset without citation was constructed by the benchmark itself.}
\resizebox{0.75\textwidth}{!}{%
\begin{tabular}{cc}
\toprule
 \textbf{Benchmark} &\textbf{Dataset}\\ \hline
 RAGAs \cite{RAGAS} &WikiEval \\\hline
 RECALL \cite{RECALL} &EventKG   \cite{Gottschalk2018}, UJ \cite{UJ}\\\hline
 ARES \cite{ARES} &\begin{tabular}[c]{@{}c@{}}NQ \cite{NQ}, Hotpot   \cite{HotpotQA},\\      FEVER \cite{FEVER}, WoW \cite{WoW}, \\      MultiRC \cite{MultiRC}, ReCoRD \cite{ReCoRD}\end{tabular}\\\hline
 RGB \cite{RGB} &Generated (Source: News)\\\hline
 MultiHop-RAG \cite{MultiHop-RAG} &Generated (Source: News)\\\hline
 CRUD-RAG \cite{CRUD} &\begin{tabular}[c]{@{}c@{}}Generated (Source: News) \\ UHGEval \cite{UHGEval}\end{tabular}\\\hline
 MedRAG \cite{MedRAGBench} &MIRAGE\\\hline
  FeB4RAG \cite{FeB4RAG}&FeB4RAG, BEIR \cite{BEIR}\\\hline
 CDQA \cite{CDQA} &Generation (Source: News), Labeller\\\hline
 DomainRAG \cite{DomainRAG}& Generation (Source: College Admission Information) \\\hline
 ReEval \cite{ReEval}&RealTimeQA\cite{RealTime}, NQ \cite{MRQA,NQ}) \\\hline
\end{tabular}
}
\label{table:dataset}
\end{table}

\subsection{Evaluation Dataset (\textit{How to evaluate?})}

In Table \ref{table:dataset}, distinct benchmarks employ varying strategies for dataset construction, ranging from leveraging existing resources to generating entirely new data tailored for specific evaluation aspects. Several benchmarks draw upon the part of KILT (Knowledge Intensive Language Tasks) benchmark \cite{KILT} (Natural Questions \cite{NQ}, HotpotQA \cite{HotpotQA}, and FEVER \cite{FEVER}) and other established datasets such as SuperGLUE \cite{SuperGLUE} (MultiRC \cite{MultiRC}, and ReCoRD \cite{ReCoRD}) \cite{ARES}. However, the drawback of using such datasets can't solve the challenges in dynamic real-world scenarios. A similar situation can be observed in WikiEval, from Wikipedia pages post 2022, constructed by RAGAs \cite{RAGAS}.

The advent of powerful LLMs has revolutionized the process of dataset construction. With the ability to design queries and ground truths for specific evaluation targets using these frameworks, authors can now create datasets in the desired format with ease. Benchmarks like RGB, MultiHop-RAG, CRUD-RAG, and CDQA \cite{RGB,MultiHop-RAG,CRUD,CDQA} have taken this approach further by building their own datasets using online news articles to test RAG systems' ability to handle real-world information beyond the training data of LM frameworks.
Most recently, DomainRAG \cite{DomainRAG} combines various types of QA datasets with single-doc, multi-doc, single-round, and multi-round. These datasets are generated from the yearly changed information from the college website for admission and enrollment, which forces the LLMs to use the provided and updated information.

In summary, the creation and selection of datasets are crucial for evaluating RAG systems. Datasets tailored for specific metrics or tasks improve evaluation accuracy and guide the development of adaptable RAG systems for real-world information needs.


\subsection{Evaluation Metric (\textit{How to quantify?})}
Navigating the intricate terrain of evaluating RAG systems necessitates a nuanced understanding of the metrics that can precisely quantify the evaluation targets. 
However, creating evaluative criteria that align with human preferences and address practical considerations is challenging. Each component within the RAG systems requires a tailored evaluative approach that reflects its distinct functionalities and objectives.

\subsubsection{Retrieval Metrics}
Various targets can be evaluated with various metrics that correspond to the given datasets. This section will introduce several commonly used metrics for retrieval and generation targets. The metrics for additional requirements can also be found in these commonly used metrics. The more specifically designed metrics can be explored in the original paper via Table \ref{table:aspect-metrics} as a reference.

For the retrieval evaluation, the focus is on metrics that can accurately capture the relevance, accuracy, diversity, and robustness of the information retrieved in response to queries. These metrics must not only reflect the system's precision in fetching pertinent information but also its resilience in navigating the dynamic, vast, and sometimes misleading landscape of available data. The deployment of metrics like \textit{Misleading Rate}, \textit{Mistake Reappearance Rate}, and \textit{Error Detection Rate} within the \cite{RECALL} benchmark underscores a heightened awareness of RAG systems' inherent intricacies. The integration of \textit{MAP@K}, \textit{MRR@K}, and \textit{Tokenization with F1} into benchmarks like \cite{MultiHop-RAG,CDQA} mirrors a deepening comprehension of traditional retrieval's multifaceted evaluation. While the \cite{EvalGenAdHocIR} also emphasizes that this ranking-based evaluation methodology is not unsuitable for the RAG system, and should have more RAG-specific retrieval evaluation metrics. These metrics not only capture the precision and recall of retrieval systems but also account for the diversity and relevance of retrieved documents, aligning with the complex and dynamic nature of information needs in RAG systems. The introduction of LLMs as evaluative judges, as seen in \cite{RAGAS}, further underscores the adaptability and versatility of retrieval evaluation, offering a comprehensive and context-aware approach to assessing retrieval quality.\\

\noindent {\textit{Non-Rank Based Metrics}} often assess binary outcomes—whether an item is relevant or not—without considering the position of the item in a ranked list. Notice, that the following formula is just one format of these metrics, the definition of each metric may vary by the different evaluating tasks.

\begin{itemize}
    \item[-] \textbf{Accuracy} is the proportion of true results (both true positives and true negatives) among the total number of cases examined.
    
    \item[-] \textbf{Precision} is the fraction of relevant instances among the retrieved instances, 
    \[
        \text{Precision} = \frac{TP}{TP + FP}
    \]
    where \(TP\) represents true positives and \(FP\) represents false positives.
    
    \item[-] \textbf{Recall} at k (\(Recall@k\)) is the fraction of relevant instances that have been retrieved over the total amount of relevant cases, considering only the top \(k\) results.
    \[
        Recall@k = \frac{|{RD} \cap {Top_{kd}}|}{|{RD}|}
    \]
    where $RD$ is the relevant documents, and $Top_{kd}$ is the top-k retrieved documents.
\end{itemize}

\noindent{\textit{Rank-Based Metrics}} evaluate the order in which relevant items are presented, with higher importance placed on the positioning of relevant items at the ranking list.

\begin{itemize}
    \item[-] \textbf{Mean Reciprocal Rank (MRR)} is the average of the reciprocal ranks of the first correct answer for a set of queries.
    \[
        MRR = \frac{1}{|Q|} \sum_{i=1}^{|Q|} \frac{1}{rank_i}
    \]
    where \(|Q|\) is the number of queries and \(rank_i\) is the rank position of the first relevant document for the \(i\)-th query.
    
    
    
    \item[-] \textbf{Mean Average Precision (MAP)} is the mean of the average precision scores for each query.
    \[
        MAP = \frac{1}{|Q|}\sum_{q=1}^{|Q|} \frac{\sum_{k=1}^{n} (P(k) \times rel(k))}{|{\text{relevant documents}}_q|}
    \]
    where \(P(k)\) is the precision at cutoff \(k\) in the list, \(rel(k)\) is an indicator function equaling 1 if the item at rank \(k\) is a relevant document, \(0\) otherwise, and \(n\) is the number of retrieved documents.
\end{itemize}

\subsubsection{Generation Metrics}
In the realm of generation, evaluation transcends the mere accuracy of generated responses, venturing into the quality of text in terms of coherence, relevance, fluency, and alignment with human judgment. This necessitates metrics that can assess the nuanced aspects of language production, including factual correctness, readability, and user satisfaction with the generated content. The traditional metrics like \textit{BLEU}, \textit{ROUGE}, and \textit{F1 Score} continue to play a crucial role, emphasizing the significance of precision and recall in determining response quality. Yet, the advent of metrics such as \textit{Misleading Rate}, \textit{Mistake Reappearance Rate}, and \textit{Error Detection Rate} highlights an evolving understanding of RAG systems' distinct challenges \cite{RECALL}.

The evaluation done by humans is still a very significant standard to compare the performance of generation models with one another or with the ground truth. The approach of employing LLMs as evaluative judges \cite{LLMasJudge} is a versatile and automatic method for quality assessment, catering to instances where traditional ground truths may be elusive \cite{RAGAS}. This methodology benefits from employing prediction-powered inference (PPI) and context relevance scoring, offering a nuanced lens through which LLM output can be assessed. \cite{ARES} The strategic use of detailed prompt templates ensures a guided assessment aligned with human preferences, effectively standardizing evaluations across various content dimensions \cite{RAGorFT}. This shift towards leveraging LLMs as arbiters mark a significant progression towards automated and context-responsive evaluation frameworks, enriching the evaluation landscape with minimal reliance on reference comparisons. 


\begin{itemize}
    \item[-] \textbf{ROUGE} Recall-Oriented Understudy for Gisting Evaluation (ROUGE) \cite{Lin2004} is a set of metrics designed to evaluate the quality of summaries by comparing them to human-generated reference summaries. ROUGE can be indicative of the content overlap between the generated text and the reference text. The variants of ROUGEs measure the overlap of n-grams (ROUGE-N, ROUGGE-W), word subsequences (ROUGE-L, ROUGGE-S), and word pairs between the system-generated summary and the reference summaries.

    \item[-] \textbf{BLEU} Bilingual Evaluation Understudy (BLEU) \cite{Papineni2002} is a metric for evaluating the quality of machine-translated text against one or more reference translations. BLEU calculates the precision of n-grams in the generated text compared to the reference text and then applies a brevity penalty to discourage overly short translations. BLEU has limitations, such as not accounting for the fluency or grammaticality of the generated text.

    \item[-] \textbf{BertScore} BertScore \cite{Zhang2020} leverages the contextual embedding from pre-trained transformers like BERT to evaluate the semantic similarity between generated text and reference text. BertScore computes token-level similarity using contextual embedding and produces precision, recall, and F1 scores. Unlike n-gram-based metrics, BertScore captures the meaning of words in context, making it more robust to paraphrasing and more sensitive to semantic equivalence.

    \item[-] \textbf{LLM as a Judge} Using ``LLM as a Judge'' for evaluating generated text is a more recent approach. \cite{LLMasJudge} In this method, LLMs are used to score the generated text based on criteria such as coherence, relevance, and fluency. The LLM can be optionally finetuned on human judgments to predict the quality of unseen text or used to generate evaluations in a zero-shot or few-shot setting. This approach leverages the LLM's understanding of language and context to provide a more nuanced text quality assessment. For instance, \cite{RAGorFT} illustrates how providing LLM judges with detailed scoring guidelines, such as a scale from 1 to 5, can standardize the evaluation process. This methodology encompasses critical aspects of content assessment, including coherence, relevance, fluency, coverage, diversity, and detail - both in the context of answer evaluation and query formulation.
\end{itemize}


\subsubsection{Additional Requirements}
These additional requirements, such as latency, diversity, noise robustness, negative rejection, and counterfactual robustness, are used to ensure the practical applicability of RAG systems in real-world scenarios aligned with human preference. This section delves into the metrics used for evaluating these additional requirements, highlighting their significance in the comprehensive assessment of RAG systems.

\textbf{Latency} measures the time taken by the RAG system to finish the response of one query. It is a critical factor for user experience, especially in interactive applications such as chatbots or search engines \cite{EfficientRAG}. \noindent\textit{Single Query Latency}: The mean time is taken to process a single query, including both retrieval and generating phases.

\textbf{Diversity} evaluates the variety and breadth of information retrieved and generated by the RAG system. It ensures that the system can provide a wide range of perspectives and avoid redundancy in responses \cite{HaystackDiversity}. \textit{Cosine Similarity / Cosine Distance}: The cosine similarity/distance calculates embeddings of retrieved documents or generated responses. \cite{Lahitani2016} Lower cosine similarity scores indicate higher diversity, suggesting that the system can retrieve or generate a broader spectrum of information.

\textbf{Noise Robustness} measures the RAG system's ability to handle irrelevant or misleading information without compromising the quality of the response \cite{RECALL}. The metrics \textit{Misleading Rate} and \textit{Mistake Reappearance Rate} are described in \cite{RECALL}, providing detailed descriptions tailored to the specific dataset and experimental setup. \cite{DomainRAG}

\textbf{Negative Rejection} evaluates the system's capability to withhold responses when the available information is insufficient or too ambiguous to provide an accurate answer \cite{RGB}. \textit{Rejection Rate}: The rate at which the system refrains from generating a response.

\textbf{Counterfactual Robustness}
Counterfactual robustness assesses the system's ability to identify and disregard incorrect or counterfactual information within the retrieved documents \cite{CRUD}. \textit{Error Detection Rate}: The ratio of counterfactual statements detected in retrieved information.

\section{\modify{Discussion}}
For RAG systems, traditional Question Answering (QA) datasets and metrics remain a common format for interaction. \cite{RAGAS,ARES,RECALL,RGB,MedRAGBench,CDQA,DomainRAG,ReEval} While these provide a basic verification of RAG's capabilities, it becomes challenging to distinguish the impact of retrieval components when faced with strong Language Models (LLMs) capable of excelling in QA benchmarks.
To comprehensively evaluate the performance of entire RAG systems, there is a need for diverse and RAG-specific benchmarks. Several papers offer guidance on improving QA format benchmarks, including variations in question types: from simple Wikipedia filling questions to multi-hop \cite{MultiHop-RAG}, multi-document questions \cite{ReEval} and single-round to multi-round dialogue \cite{CRUD,DomainRAG}. For answers, aspects such as structural output \cite{DomainRAG}, content moderation \cite{RGB,TruEra}, and hallucination \cite{ReEval} can be considered when evaluating relevance, faithfulness, and correctness. In addition to these, RAG systems require additional requirements such as robustness to noisy documents, language expression, latency, and result diversity. \cite{LangChain2023,DatabricksRAGEval,RECALL,RGB,CRUD,FeB4RAG,DomainRAG,EfficientRAG,HaystackDiversity} Furthermore, research is needed on performance changes involving intermediate outputs and retrieved documents, as well as the relationship and analysis between retrieval metrics and final generation outputs.

Regarding \textit{datasets}, creating a universal dataset was challenging due to the target-specific nature of different RAG benchmarks. Tailored datasets  \cite{RAGAS,RECALL,ARES,CRUD,FeB4RAG} are necessary for a thorough evaluation, but this approach increases the effort and resources required. Moreover, the diversity of datasets, from news articles to structured databases \cite{ReEval}, reflects the adaptability required of RAG systems but also poses a barrier to streamlined evaluation. Recently, with the cutting-edge performance of LLMs, complex data processing and automatic QA pair generation can be automated to achieve daily or finer-grained time resolution, preventing LLMs from cheating and evaluating the robustness of RAG systems in rapidly changing data. \cite{RGB,MultiHop-RAG,CRUD,CDQA,DomainRAG,ReEval} 

When it comes to \textit{metrics}, the use of LLMs as automatic evaluative judges signifies a burgeoning trend, promising versatility and depth in generative outputs with reasoning on a large scale compared to human evaluation. However, using ``LLMs as a Judge'' \cite{LLMasJudge} for responses presents challenges in aligning with human judgment, establishing effective grading scales, and applying consistent evaluation across varied use cases. Determining correctness, clarity, and richness can differ between automated and human assessments. Moreover, the effectiveness of example-based scoring can vary, and there's no universally applicable grading scale and prompting text, complicating the standardization of ``LLM as a Judge''. \cite{DatabricksRAGEval}

In addition to the challenges mentioned above, it is important to consider the resource-intensive nature \cite{Green} of using Large Language Models (LLMs) for data generation and validation. RAG benchmarks must balance the need for thorough evaluation with the practical constraints of limited computational resources. As such, it is desirable to develop evaluation methodologies that can effectively assess RAG systems using smaller amounts of data while maintaining the validity and reliability of the results.

\section{\modify{Conclusion}}
This survey systematically explores the complexities of evaluating RAG systems, highlighting the challenges in assessing their performance. Through the proposed \textit{\lname{}}, we outline a structured approach to analyzing RAG evaluations, focusing on targets, datasets and measures. Our analysis emphasizes the need for targeted benchmarks that reflect the dynamic interplay between retrieval accuracy and generative quality and practical considerations for real-world applications.
By identifying gaps in current methodologies and suggesting future research directions, we aim to contribute to more effective, and user-aligned benchmarks of RAG systems.



\newpage
\bibliographystyle{splncs04}
\bibliography{custom}

\appendix
\newpage

\section{Structure of RAG System}
\label{appendix:rag-structure}

\subsection{Retrieval Component}
The retrieval component of RAG systems in Figure \ref{fig:rag-structure} can be categorized into three types: sparse retrieval, dense retrieval \cite{Zhu2021}, and web search engine. The standard for evaluation is the output of \textit{relevant documents} with numerical scores or rankings. 

Before the introduction of neural networks, \textit{sparse retrievals} are widely used for retrieving relative text content. Methods like TF-IDF \cite{Ramos2003} and BM25 \cite{Robertson2009} rely on keyword matching and word frequency but may miss semantically relevant documents without keyword overlap.


By leveraging deep learning models such as BERT \cite{Devlin2019}, \textit{dense retrieval} can capture the semantic meaning of texts, which allows them to find relevant documents even when keyword overlap is minimal. This is crucial for complex queries that require a contextual understanding to retrieve accurate information. With advanced fusion structure for queries and documents \cite{ColBERT} and the more efficient implementation of K-Nearest Neighbors (KNN) \cite{Shahabi2002}, Approximate Nearest Neighbor (ANN) \cite{FAISS,Johnson2019} search techniques, dense retrieval methods have become practical for large-scale use. 

\textit{Web search engine} employs the complex online search engine to provide relevant documents, such as Google Search \cite{GoogleSearch}, Bing Search \cite{MicrosoftBing}, DuckDuckGo \cite{DuckDuckGo}. RAG systems can traverse the web's extensive information, potentially returning a more diverse and semantically relevant set of documents via the API of the search provider. The black box of the search engine and the expense of large-scale search are not affordable sometimes.

It is observed that dense retrieval techniques, particularly those leveraging embeddings, stand out as the preferred choice within the RAG ecosystem. These methods are frequently employed in tandem with sparse retrieval strategies, creating a hybrid approach that balances precision and breadth in information retrieval. Moreover, the adoption of sophisticated web search engines for benchmark assessment underscores their growing significance in enhancing the robustness and comprehensiveness of evaluations.

\subsubsection{Indexing}
The indexing component processes and indexes document collections, such as HuggingFace datasets or Wikipedia pages. Chunking before indexing can improve retrieval by limiting similarity scores to individual chunks, as semantic embedding is less accurate for long articles, and desired content is often brief \cite{LangChain2023}. Index creation is designed for fast and efficient search. For example, the inverted index for sparse retrieval and the ANN index for dense retrieval.

\textit{Sparse Retrieval} involves calculating IDF for each term and storing values in a database for quick look-up and scoring when queried.

\textit{Dense Retrieval} encodes documents into dense vectors using a pre-trained language model like BERT. These vectors are then indexed using an Approximate Nearest Neighbor (ANN) search technique, like graph-based Hierarchical Navigable Small World (HNSW) or Inverted File Index (IVF) \cite{FAISS}. This process allows for the efficient retrieval of ``closed'' items by given predefined distance metrics.

\subsubsection{Search}
This step is responsible for retrieving relevant documents based on a given query. Queries are submitted using the respective API to retrieve relevant documents for web search engine retrieval. For local resources, the query component is responsible for formatting the query in the format required by different sparse or dense retrieval methods. Then, the query is submitted to the retrieval system, which returns a set of relevant documents along with their scores.

In both local and web-based scenarios, an optional reranker can be employed to refine the ranking of retrieved documents further. The reranker usually comprises a more complex and larger model that considers additional features of the documents and the given query. These additional features often include the semantic relationship between the query and the document content, document importance or popularity, and other custom measures specific to the information need at hand.


\subsection{Generation Component}
The evaluable output for the generation component is the \textit{response} of LLMs and the \textit{structured or formatted output} from the phrased response.

\subsubsection{Prompting}
The generation process critically hinges on prompting, where a query, retrieval outcomes, and instructions converge into a single input for the language model. Research showcases various strategic prompting tactics such as the Chain of Thought (CoT) \cite{CoT}, Tree of Thought (ToT) \cite{ToT}, and Self-Note \cite{SelfNote}, each significantly shaping the model's output. These methods, especially the step-by-step approach, are pivotal in augmenting LLMs for intricate tasks.

Prompting innovations have introduced methods like Rephrase and Respond (RaR) \cite{RaR}, enhancing LLMs by refining queries within prompts for better comprehension and response. This technique has proven to boost performance across diverse tasks.
The latest RAG benchmarks \cite{MedRAGBench,CDQA} in the specific domains start to evaluate the robustness of various prompting engineering skills, including CoT, RaR, etc.

\subsubsection{Inference}

The final input string prepared in the prompting step is then passed on to the LLMs as input, which generates the output. The inference stage is where the LLM operates on the input derived from the retrieval and the prompting stages in the pipeline to generate the final output. This is usually the answer to the initial query and is used for downstream tasks.

Depending on the specifics of the task or expected output structure, a post-processing step may be implemented here to format the generated output suitably or extract specific information from the response. For example, the classification problems (multichoice questions) or if the task requires the extraction of specific information from the generated text, this step could involve additional named entity recognition or parsing operations.

\end{document}